\documentclass{article}

\usepackage{here}

\usepackage{times}
\usepackage{graphicx} 
\usepackage{subfigure} 
\usepackage{epstopdf}
\usepackage{natbib}

\usepackage{algorithm}
\usepackage{algorithmic}

\usepackage{hyperref}

\usepackage[accepted]{icml-like} 
\usepackage{filecontents}
\usepackage{amssymb}
\usepackage{amsmath}
\icmltitlerunning{Convolutional Neural Networks using Logarithmic Data Representation}

\begin{document} 

\twocolumn[
\icmltitle{Convolutional Neural Networks using Logarithmic Data Representation}

\icmlauthor{Daisuke Miyashita}{daisukem@stanford.edu}
\icmladdress{Stanford University,
            Stanford, CA 94305 USA \\
            Toshiba, Kawasaki, Japan}
\icmlauthor{Edward H. Lee}{edhlee@stanford.edu}
\icmladdress{Stanford University,
            Stanford, CA 94305 USA}
\icmlauthor{Boris Murmann}{murmann@stanford.edu}
\icmladdress{Stanford University,
            Stanford, CA 94305 USA}

\icmlkeywords{deep learning, convolutional neural network, machine learning}

\vskip 0.3in
]

\begin{abstract}
Recent advances in convolutional neural networks have considered model complexity and hardware efficiency to enable deployment onto embedded systems and mobile devices. For example, it is now well-known that the arithmetic operations of deep networks can be encoded down to 8-bit fixed-point without significant deterioration in performance. However, further reduction in precision down to as low as $3$-bit fixed-point results in significant losses in performance. In this paper we propose a new data representation that enables state-of-the-art networks to be encoded to 3 bits with negligible loss in classification performance. To perform this, we take advantage of the fact that the weights and activations in a trained network naturally have non-uniform distributions. Using non-uniform, base-2 logarithmic representation to encode weights, communicate activations, and perform dot-products enables networks to $1$) achieve higher classification accuracies than fixed-point at the same resolution and $2$) eliminate bulky digital multipliers. Finally, we propose an end-to-end training procedure that uses $\log$ representation at $5$-bits, which achieves higher final test accuracy than linear at $5$-bits.

\end{abstract} 

\section{Introduction}
Deep convolutional neural networks (CNN) have demonstrated state-of-the-art performance in image classification \cite{Krizhevsky12,vgg,He15} but have steadily grown in computational complexity. For example, the Deep Residual Learning \cite{He15} set a new record in image classification accuracy at the expense of $11.3$ billion floating-point multiply-and-add operations per forward-pass of an image and $230$ MB of memory to store the weights in its $152$-layer network. 

In order for these large networks to run in real-time applications such as for mobile or embedded platforms, it is often necessary to use low-precision arithmetic and apply compression techniques. Recently, many researchers have successfully deployed networks that compute using $8$-bit fixed-point representation \cite{Vanhoucke11, tensorflow2015-whitepaper} and have successfully trained networks with $16$-bit fixed point \cite{Gupta15}. This work in particular is built upon the idea that algorithm-level noise tolerance of the network can motivate simplifications in hardware complexity.

Interesting directions point towards matrix factorization \cite{Denton14} and tensorification \cite{Novikov15} by leveraging structure of the fully-connected (FC) layers. Another promising area is to prune the FC layer before mapping this to sparse matrix-matrix routines in GPUs \cite{NIPS_Song}. However, many of these inventions aim at systems that meet some required and specific criteria such as networks that have many, large FC layers or accelerators that handle efficient sparse matrix-matrix arithmetic. And with network architectures currently pushing towards increasing the depth of convolutional layers by settling for fewer dense FC layers \cite{He15, Szegedy15}, there are potential problems in motivating a one-size-fits-all solution to handle these computational and memory demands.

We propose a general method of representing and computing the dot products in a network that can allow networks with minimal constraint on the layer properties to run more efficiently in digital hardware. In this paper we explore the use of communicating activations, storing weights, and computing the atomic dot-products in the binary logarithmic (base-2 logarithmic) domain for both inference and training. The motivations for moving to this domain are the following:

\begin{itemize}
\item Training networks with weight decay leads to final weights that are distributed non-uniformly around $0$. 
\item Similarly, activations are also highly concentrated near $0$. Our work uses rectified Linear Units (ReLU) as the non-linearity.  
\item  Logarithmic representations can encode data with very large dynamic range in fewer bits than can fixed-point representation \cite{Gautschi16}. 
\item Data representation in $\log$-domain is naturally encoded in digital hardware (as shown in Section \ref{sec:log quant of conv}).
\end{itemize}

Our contributions are listed:
\begin{itemize}
\item we show that networks obtain higher classification accuracies with logarithmic quantization than linear quantization using traditional fixed-point at equivalent resolutions.
\item we show that activations are more robust to quantization than weights. This is because the number of activations tend to be larger than the number of weights which are reused during convolutions.
\item we apply our logarithmic data representation on state-of-the-art networks, allowing activations and weights to use only $3$b with almost no loss in classification performance. 
\item we generalize base-$2$ arithmetic to handle different base. In particular, we show that a base-$\sqrt{2}$ enables the ability to capture large dynamic ranges of weights and activations but also finer precisions across the encoded range of values as well.
\item we develop logarithmic backpropagation for efficient training.
\end{itemize}

\section{Related work}
\textbf{Reduced-precision computation.}
\cite{Shin15, Sung15, Vanhoucke11, ICLR_Song15} analyzed the effects of quantizing the trained weights for inference. For example, \cite{NIPS_Song} shows that convolutional layers in AlexNet \cite{Krizhevsky12} can be encoded to as little as 5 bits without a significant accuracy penalty. There has also been recent work in training using low precision arithmetic. \cite{Gupta15} propose a stochastic rounding scheme to help train networks using 16-bit fixed-point. \cite{Lin15} propose quantized back-propagation and ternary connect. This method reduces the number of floating-point multiplications by casting these operations into powers-of-two multiplies, which are easily realized with bitshifts in digital hardware. They apply this technique on MNIST and CIFAR10 with little loss in performance. However, their method does not completely eliminate all multiplications end-to-end. During test-time the network uses the learned full resolution weights for forward propagation. Training with reduced precision is motivated by the idea that high-precision gradient updates is unnecessary for the stochastic optimization of networks \cite{Bottou07, Bishop95, Audhkhasi13}. In fact, there are some studies that show that gradient noise helps convergence. For example, \cite{Neelakantan15} empirically finds that gradient noise can also encourage faster exploration and annealing of optimization space, which can help network generalization performance.

\textbf{Hardware implementations.}
There have been a few but significant advances in the development of specialized hardware of large networks. For example \cite{Farabet10} developed Field-Programmable Gate Arrays (FPGA) to perform real-time forward propagation. These groups have also performed a comprehensive study of classification performance and energy efficiency as function of resolution. \cite{Zhang15} have also explored the design of convolutions in the context of memory versus compute management under the RoofLine model. Other works focus on specialized, optimized kernels for general purpose GPUs \cite{cuDNN}. 


\section{Concept and Motivation}
\label{sssec:concept}

\begin{figure}[ht]
\vskip 0.2in
\begin{center}
\centerline{\includegraphics[width=\columnwidth]{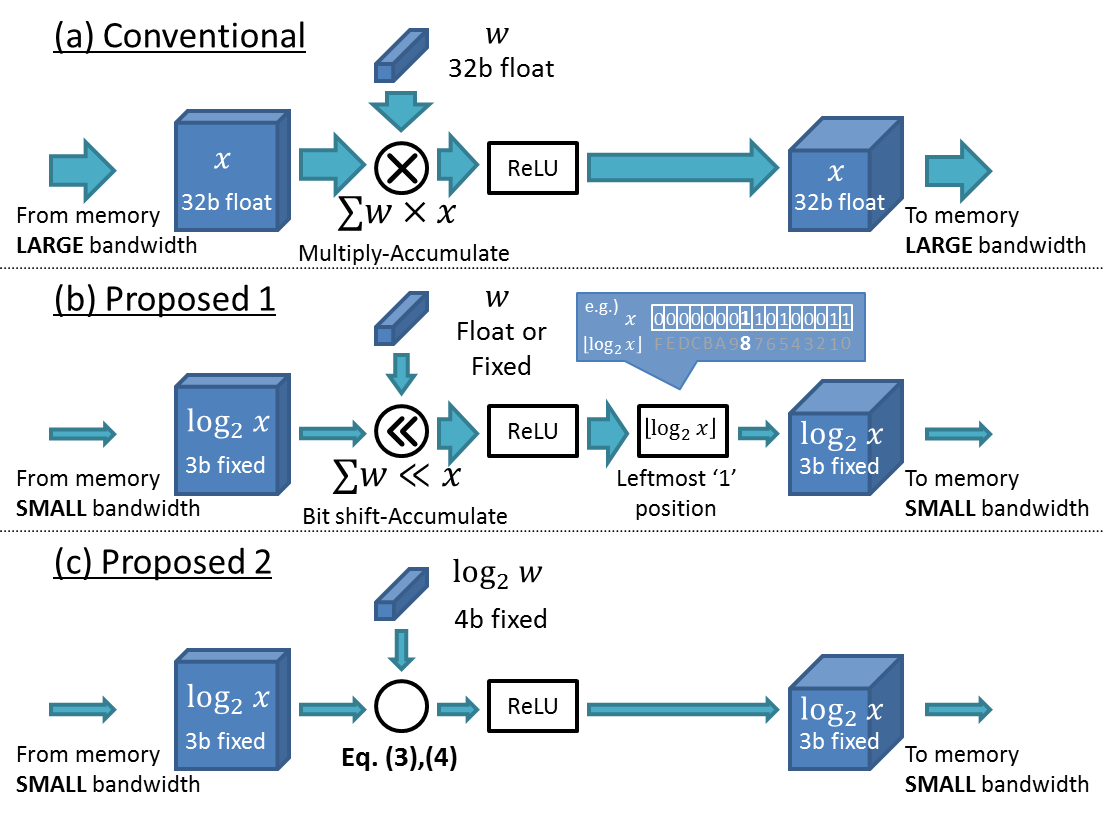}}
\caption{Concept and motivation of this study.}
\label{fig:concept}
\end{center}
\vskip -0.2in
\end{figure} 
Each convolutional and fully-connected layer of a network performs matrix operations that distills down to dot products $y=w^Tx$, where $x \in \mathbb{R}^n$ is the input, $w \in \mathbb{R}^n$ the weights, and $y$ the activations before being transformed by the non-linearity (e.g. ReLU). Using conventional digital hardware, this operation is performed using $n$ multiply-and-add operations using floating or fixed point representation as shown in Figure~\ref{fig:concept}(a). However, this dot product can also be computed in the $\log$-domain as shown in Figure~\ref{fig:concept}(b,c).

\subsection{Proposed Method 1.}
\label{ssec:sub1}
The first proposed method as shown in Figure~\ref{fig:concept}(b) is to transform one operand to its $\log$ representation, convert the resulting transformation back to the linear domain, and multiply this by the other operand. This is simply 
\begin{eqnarray}
\label{eq:xonly}
w^T x & \simeq & \sum_{i=1}^{n} w_i \times 2^{\tilde{x}_i} \nonumber \\
& = & \sum_{i=1}^{n} \mathrm{Bitshift}(w_i, \tilde{x}_i),
\end{eqnarray}
where $\tilde{x}_i =  \mathrm{Quantize}(\log _{2}(x_i))$, $\mathrm{Quantize}(\bullet)$ quantizes $\bullet$ to an integer, and $\mathrm{Bitshift}(a, b)$ is the function that bitshifts a value $a$ by an integer $b$ in fixed-point arithmetic. In floating-point, this operation is simply an addition of $b$ with the exponent part of $a$. Taking advantage of the  $\mathrm{Bitshift}(a, b)$ operator to perform multiplication obviates the need for expensive digital multipliers.

Quantizing the activations and weights in the $\log$-domain ($\log_2(x)$ and $\log_2(w)$) instead of $x$ and $w$ is also motivated by leveraging structure of the non-uniform distributions of $x$ and $w$. A detailed treatment is shown in the next section. In order to quantize, we propose two hardware-friendly flavors. The first option is to simply floor the input. This method computes $ \lfloor \log _{2}(w) \rfloor $ by returning the position of the first $1$ bit seen from the most significant bit (MSB). The second option is to round to the nearest integer, which is more precise than the first option. With the latter option, after computing the integer part, the fractional part is computed in order to assert the rounding direction. This method of rounding is summarized as follows. Pick $m$ bits followed by the leftmost $1$ and consider it as a fixed point number $F$ with 0 integer bit and $m$ fractional bits. Then, if $F \geq \sqrt{2}-1$, round $F$ up to the nearest integer and otherwise round it down to the nearest integer.

\subsection{Proposed Method 2.}
\label{ssec:sub2}
The second proposed method as shown in Figure~\ref{fig:concept}(c) is to extend the first method to compute dot products in the $\log$-domain for both operands.
Additions in linear-domain map to sums of exponentials in the $\log$-domain and multiplications in linear become $\log$-addition. The resulting dot-product is

\begin{eqnarray}
\label{eq:wandx}
w^Tx & \simeq &  \sum_{i=1}^{n} 2 ^{\mathrm{Quantize}(\log _{2}(w_i))+\mathrm{Quantize}(\log _{2}(x_i))}  \nonumber \\
& = & \sum_{i=1}^{n} \mathrm{Bitshift}(1,\tilde{w}_i + \tilde{x}_i),
\end{eqnarray}
 
where the $\log$-domain weights are $\tilde{w}_i= \mathrm{Quantize}(\log _{2}(w_i))$ and $\log$-domain inputs are $\tilde{x}_i = \mathrm{Quantize}(\log _{2}(x_i))$.

By transforming both the weights and inputs, we compute the original dot product by bitshifting $1$ by an integer result $\tilde{w}_i + \tilde{x}_i$ and summing over all $i$.

\subsection{Accumulation in $\log$ domain}
\label{sec:log acc}
Although Fig.~\ref{fig:concept}(b,c) indicates a logarithm-to-linear converter between layers where the actual accumulation is performed in the linear domain, this accumulation is able to be performed in the $\log$-domain using the approximation $ \log _{2}(1+x) \simeq x $ for $0 \leq x < 1$. For example, let $s_n = w_1x_1 + \ldots + w_nx_n$, $\tilde{s}_n = \log_2(s_n)$, and $\tilde{p}_i= \tilde{w}_i+\tilde{x}_i$. When $n=2$,
\begin{eqnarray}
\label{eq:wandx2}
\tilde{s}_2 & = & \log_{2}\left(\sum_{i=1}^{2} \mathrm{Bitshift}\left(1,\tilde{p}_i\right)\right) \nonumber \\
& \simeq & \mathrm{max} \left( \tilde{p}_{1}, \tilde{p}_{2} \right) + \mathrm{Bitshift}\left(1,-|\tilde{p}_{1}- \tilde{p}_{2}|\right),
\end{eqnarray}
and for $n$ in general,
\begin{eqnarray}
\label{eq:wandx3}
\tilde{s}_n \simeq \mathrm{max} \left( \tilde{s}_{n-1}, \tilde{p}_{n} \right) + \mathrm{Bitshift}\left(1,-|\lfloor \tilde{s}_{n-1} \rfloor - \tilde{p}_{n}| \right).
\end{eqnarray}

Note that $\tilde{s}_i$ preserves the fractional part of the word during accumulation. Both accumulation in linear domain and accumulation in $\log$ domain have its pros and cons. Accumulation in linear domain is simpler but requires larger bit widths to accommodate large dynamic range numbers. Accumulation in $\log$ in (\ref{eq:wandx2}) and (\ref{eq:wandx3}) appears to be more complicated, but is in fact simply computed using bit-wise operations in digital hardware.

\section{Experiments of Proposed Methods}
\label{sec:evaluation}
Here we evaluate our methods as detailed in Sections ~\ref{ssec:sub1} and ~\ref{ssec:sub2} on the classification task of ILSVRC-2012 \cite{imagenet_cvpr09} using Chainer ~\cite{chainer15}. We evaluate method 1 (Section ~\ref{ssec:sub1}) on inference (forward pass) in Section~\ref{sec:log quant of act}. Similarly, we evaluate method 2 (Section ~\ref{ssec:sub2}) on inference in Sections ~\ref{sec:log quant of fc} and ~\ref{sec:log quant of conv}. For those experiments, we use published models (AlexNet \cite{Krizhevsky12}, VGG16 \cite{vgg}) from the caffe model zoo (\cite{caffe}) without any fine tuning (or extra retraining). Finally, we evaluate method 2 on training in Section ~\ref{sec:log quant of delta}. 

\subsection{Logarithmic Representation of Activations}
\label{sec:log quant of act}

\begin{table}[t]
\caption{Structure of AlexNet\cite{Krizhevsky12} with quantization}
\label{table:alexnet}
\vskip 0.15in
\begin{center}
\begin{small}
\begin{tabular}{lccc}
\hline
\abovespace\belowspace
layer & \# Weight & \# Input & $\mathrm{FSR}$ \\
\hline
\abovespace
ReLU(Conv1)   & $96\cdot3\cdot11^{2}$ & $3\cdot227^{2}$ & - \\
{\bf LogQuant1} & - & $96\cdot55^{2}$ & $\mathrm{fsr}+3$ \\
LRN1 & - & - & - \\
Pool1 & - & $96\cdot55^{2}$ & - \\
ReLU(Conv2) & $256\cdot96\cdot5^{2}$ & $96\cdot27^{2}$ & - \\
{\bf LogQuant2} & - & $256\cdot27^{2}$ & $\mathrm{fsr}+3$ \\
LRN2 & - & - & - \\
Pool2 & - & $256\cdot27^{2}$ & - \\
ReLU(Conv3) & $384\cdot256\cdot3^{2}$ & $256\cdot13^{2}$ & - \\
{\bf LogQuant3} & - & $384\cdot13^{2}$ & $\mathrm{fsr}+4$ \\
ReLU(Conv4) & $384\cdot384\cdot3^{2}$ & $384\cdot13^{2}$ & - \\
{\bf LogQuant4} & - & $384\cdot13^{2}$ & $\mathrm{fsr}+3$ \\
ReLU(Conv5) & $256\cdot384\cdot3^{2}$ & $384\cdot13^{2}$ & - \\
{\bf LogQuant5} & - & $256\cdot13^{2}$ & $\mathrm{fsr}+3$ \\
Pool5 & - & $256\cdot13^{2}$ & - \\
ReLU(FC6) & $4096\cdot256\cdot6^{2}$ & $256\cdot6^{2}$ & - \\
{\bf LogQuant6} & - & $4096$ & $\mathrm{fsr}+1$ \\
ReLU(FC7) & $4096\cdot4096$ & $4096$ & - \\
{\bf LogQuant7} & - & $4096$ & $\mathrm{fsr}$ \\
\belowspace
FC8 & $1000\cdot4096$ & $4096$ & - \\
\hline
\end{tabular}
\end{small}
\end{center}
\vskip -0.1in
\end{table}

\begin{table}[t]
\caption{Structure of VGG16\cite{vgg} with quantization}
\label{table:vgg}
\vskip 0.15in
\begin{center}
\begin{small}
\begin{tabular}{lccc}
\hline
\abovespace\belowspace
layer & \# Weight & \# Input & $\mathrm{FSR}$ \\
\hline
\abovespace
ReLU(Conv1\_1)  & $64\cdot3\cdot3^{2}$  & $3\cdot224^{2}$  & - \\
{\bf LogQuant1\_1} & -                  & $64\cdot224^{2}$ & $\mathrm{fsr}+4$ \\
ReLU(Conv1\_2)  & $64\cdot64\cdot3^{2}$  & $64\cdot224^{2}$ & - \\
{\bf LogQuant1\_2} & -                  & $64\cdot224^{2}$ & $\mathrm{fsr}+6$ \\
Pool1           & -                     & $64\cdot224^{2}$ & - \\
ReLU(Conv2\_1)  & $128\cdot64\cdot3^{2}$ & $64\cdot112^{2}$  & - \\
{\bf LogQuant2\_1} & -                  & $128\cdot112^{2}$ & $\mathrm{fsr}+6$ \\
ReLU(Conv2\_2)  & $128\cdot128\cdot3^{2}$ & $128\cdot112^{2}$ & - \\
{\bf LogQuant2\_2} & -                  & $128\cdot112^{2}$ & $\mathrm{fsr}+7$ \\
Pool2           & -                     & $128\cdot112^{2}$ & - \\
ReLU(Conv3\_1)  & $256\cdot128\cdot3^{2}$ & $128\cdot56^{2}$  & - \\
{\bf LogQuant3\_1} & -                  & $256\cdot56^{2}$ & $\mathrm{fsr}+7$ \\
ReLU(Conv3\_2)  & $256\cdot256\cdot3^{2}$ & $256\cdot56^{2}$ & - \\
{\bf LogQuant3\_2} & -                  & $256\cdot56^{2}$ & $\mathrm{fsr}+7$ \\
ReLU(Conv3\_3)  & $256\cdot256\cdot3^{2}$ & $256\cdot56^{2}$ & - \\
{\bf LogQuant3\_3} & -                  & $256\cdot56^{2}$ & $\mathrm{fsr}+7$ \\
Pool3           & -                     & $256\cdot56^{2}$ & - \\
ReLU(Conv4\_1)  & $512\cdot256\cdot3^{2}$ & $256\cdot28^{2}$  & - \\
{\bf LogQuant4\_1} & -                  & $512\cdot28^{2}$ & $\mathrm{fsr}+7$ \\
ReLU(Conv4\_2)  & $512\cdot512\cdot3^{2}$ & $512\cdot28^{2}$ & - \\
{\bf LogQuant4\_2} & -                  & $512\cdot28^{2}$ & $\mathrm{fsr}+6$ \\
ReLU(Conv4\_3)  & $512\cdot512\cdot3^{2}$ & $512\cdot28^{2}$ & - \\
{\bf LogQuant4\_3} & -                  & $512\cdot28^{2}$ & $\mathrm{fsr}+5$ \\
Pool4           & -                     & $512\cdot28^{2}$ & - \\
ReLU(Conv5\_1)  & $512\cdot512\cdot3^{2}$ & $512\cdot14^{2}$  & - \\
{\bf LogQuant5\_1} & -                  & $512\cdot14^{2}$ & $\mathrm{fsr}+4$ \\
ReLU(Conv5\_2)  & $512\cdot512\cdot3^{2}$ & $512\cdot14^{2}$ & - \\
{\bf LogQuant5\_2} & -                  & $512\cdot14^{2}$ & $\mathrm{fsr}+3$ \\
ReLU(Conv5\_3)  & $512\cdot512\cdot3^{2}$ & $512\cdot14^{2}$ & - \\
{\bf LogQuant5\_3} & -                  & $512\cdot14^{2}$ & $\mathrm{fsr}+2$ \\
Pool5           & -                     & $512\cdot14^{2}$ & - \\
ReLU(FC6)    & $4096\cdot512\cdot7^{2}$ & $512\cdot7^{2}$  & - \\
{\bf LogQuant6} & -                     & $4096$           & $\mathrm{fsr}+1$ \\
ReLU(FC7)       & $4096\cdot4096$       & $4096$           & - \\
{\bf LogQuant7} & -                     & $4096$           & $\mathrm{fsr}$ \\
\belowspace
FC8             & $1000\cdot4096$       & $4096$           & - \\
\hline
\end{tabular}
\end{small}
\end{center}
\vskip -0.1in
\end{table}
This experiment evaluates the classification accuracy using logarithmic activations and floating point 32b for the weights.  In similar spirit to that of \cite{Gupta15}, we describe the logarithmic quantization layer {\bf LogQuant} that performs the element-wise operation as follows:
\begin{eqnarray}
\label{eq:logquant}
\mathrm{LogQuant}(x, \mathrm{bitwidth}, \mathrm{FSR}) & = & \left \{ \begin{array}{ll}
0 & x = 0, \\
2^{\tilde{x}} & \mathrm{otherwise,}
\end{array} \right.
\end{eqnarray}where
\begin{eqnarray}
\tilde{x} = \mathrm{Clip}\left(\mathrm{Round}(\log _{2}(|x|)), \mathrm{FSR}-2^{\mathrm{bitwidth}}, \mathrm{FSR}\right), \\
\mathrm{Clip}(x, \mathrm{min}, \mathrm{max}) = \left \{ \begin{array}{ll}
0 & x \leq \mathrm{min}, \\
\mathrm{max}-1 & x \geq \mathrm{max}, \\
x & \mathrm{otherwise.}
\end{array} \right.
\end{eqnarray}
These layers perform the logarithmic quantization and computation as detailed in Section ~\ref{ssec:sub1}. Tables~\ref{table:alexnet} and ~\ref{table:vgg} illustrate the addition of these layers to the models. 
The quantizer has a specified full scale range, and this range in linear scale is $2^{\mathrm{FSR}}$, where we express this as simply $\mathrm{FSR}$ throughout this paper for notational convenience. The $\mathrm{FSR}$ values for each layer are shown in Tables~\ref{table:alexnet} and ~\ref{table:vgg}; they show $\mathrm{fsr}$ added by an offset parameter. This offset parameter is chosen to properly handle the variation of activation ranges from layer to layer using $100$ images from the training set. The $\mathrm{fsr}$ is a parameter which is global to the network and is tuned to perform the experiments to measure the effect of $\mathrm{FSR}$ on classification accuracy. The $\mathrm{bitwidth}$ is the number of bits required to represent a number after quantization. Note that since we assume applying quantization after ReLU function, $x$ is 0 or positive and then we use unsigned format without sign bit for activations.

\begin{figure}[ht]
\vskip 0.2in
\begin{center}
\centerline{\includegraphics[width=\columnwidth]{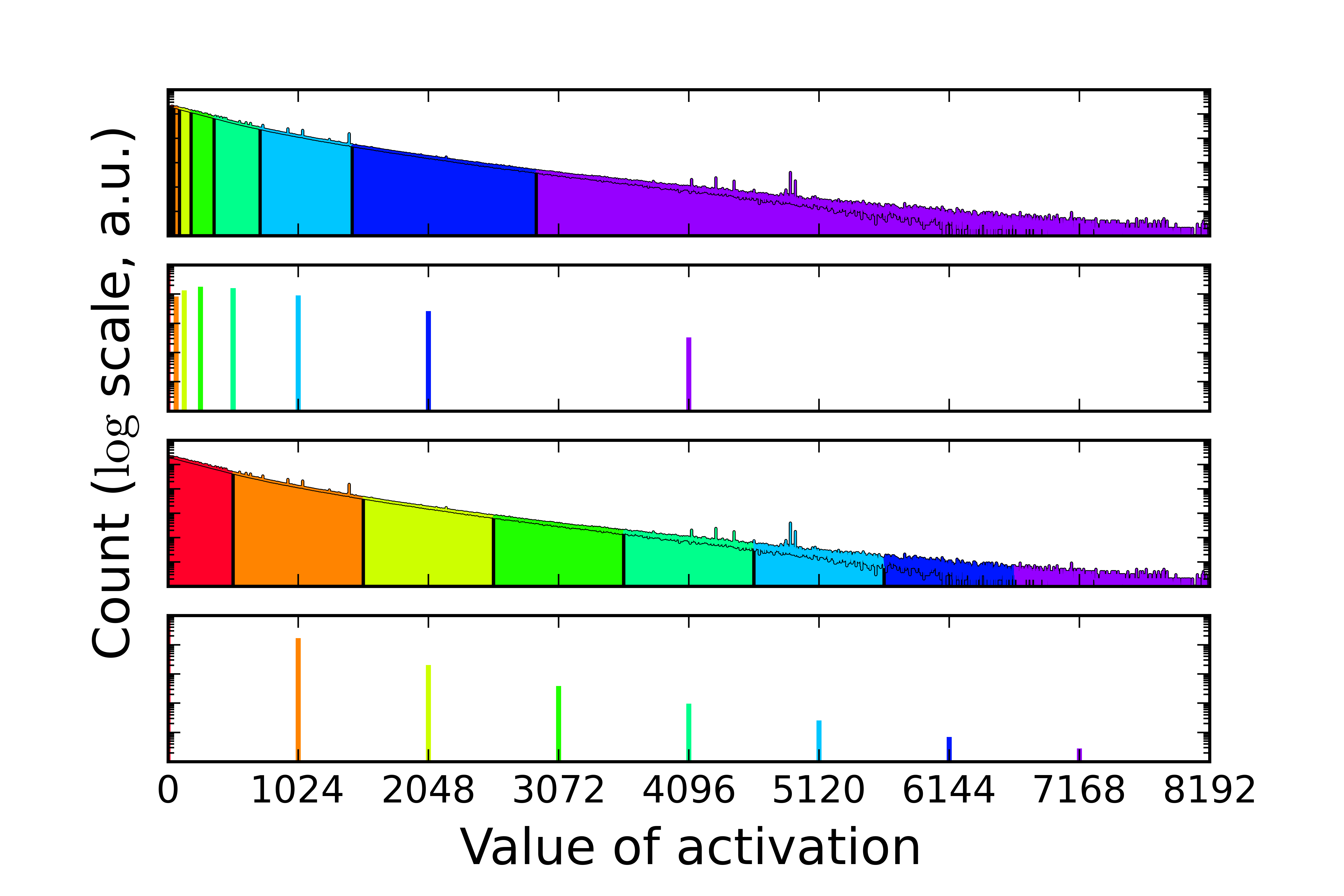}}
\caption{Distribution of activations of conv2\_2 layer in VGG16 before and after $\log$ and linear quantization. The order (from top to bottom) is: before $\log$-quantization, after $\log$-quantization, before linear quantization, and after linear quantization. The color highlights the binning process of these two quantizers.}
\label{fig:quantization}
\end{center}
\vskip -0.2in
\end{figure}

\begin{figure}[ht]
\begin{center}
\centerline{\includegraphics[width=\columnwidth]{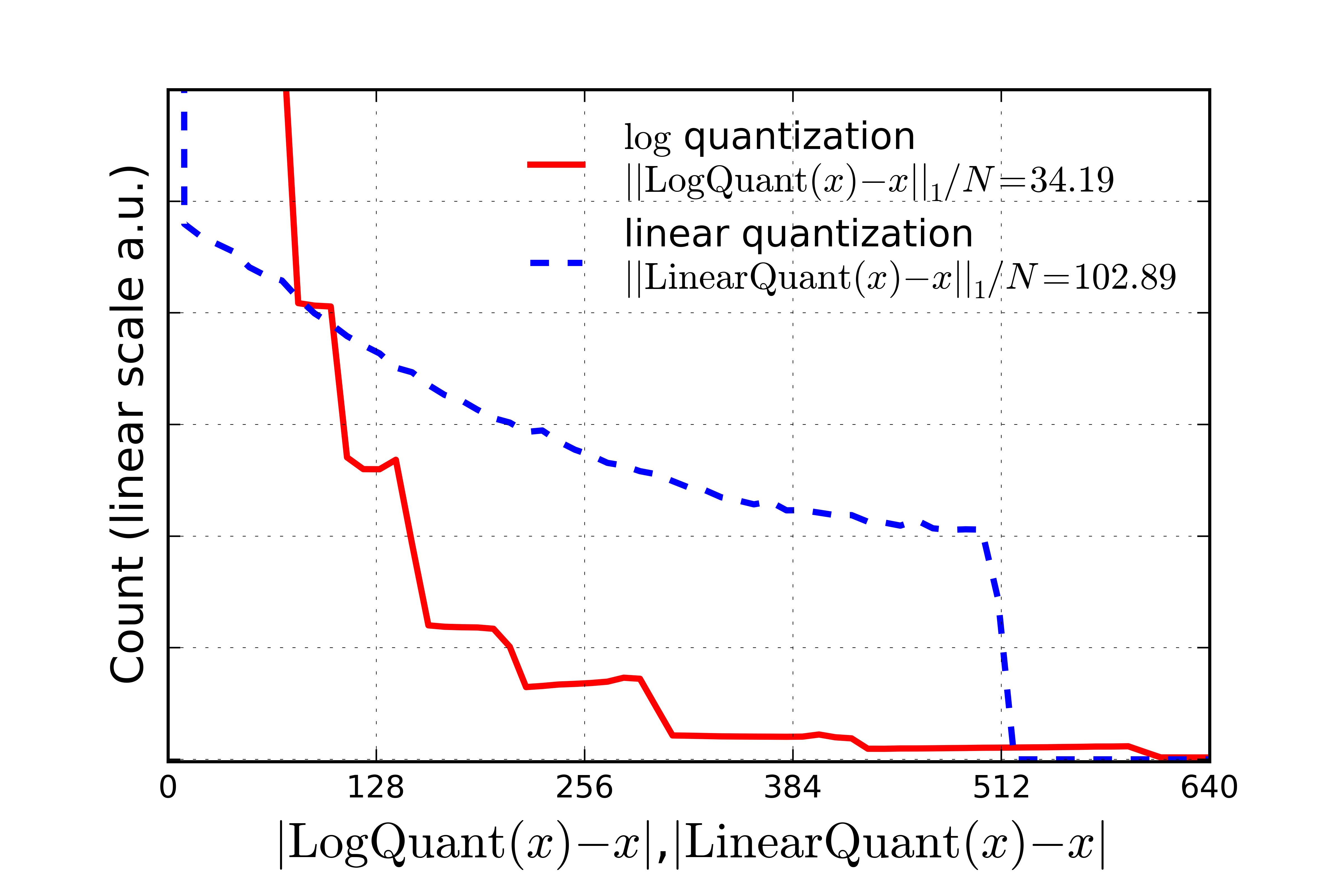}}
\caption{Comparison of the quantization error distribution between logarithmic quantization and linear quantization}
\label{fig:error act}
\end{center}
\end{figure}

In order to evaluate our logarithmic representation, we detail an equivalent linear quantization layer described as
\begin{eqnarray}
\label{eq:linquant}
\mathrm{LinearQuant}(x, \mathrm{bitwidth}, \mathrm{FSR}) \nonumber \\
= \mathrm{Clip}\left(\mathrm{Round}\left(\frac{x}{\mathrm{step}}\right)\times \mathrm{step},0,2^{\mathrm{FSR}}\right)
\end{eqnarray}
 and where 
\begin{eqnarray}
&\mathrm{step} = 2^{\mathrm{FSR}-\mathrm{bitwidth}}.
\end{eqnarray}
Figure~\ref{fig:quantization} illustrates the effect of the quantizer on activations following the conv2\_2 layer used in VGG16. The pre-quantized distribution tends to 0 exponentially, and the $\log$-quantized distribution illustrates how the $\log$-encoded activations are uniformly equalized across many output bins which is not prevalent in the linear case. Many smaller activation values are more finely represented by $\log$ quantization compared to linear quantization. The total quantization error $\frac{1}{N}||\mathrm{Quantize}(x)-x||_1$, where $\mathrm{Quantize}(\bullet)$ is $\mathrm{LogQuant}(\bullet)$ or $\mathrm{LinearQuant}(\bullet)$, $x$ is the vectorized activations of size $N$, is less for the $\log$-quantized case than for linear. This result is illustrated in Figure \ref{fig:error act}. Using linear quantization with step size of 1024, we obtain a distribution of quantization errors that are highly concentrated in the region where $|\mathrm{LinearQuant}(x)-x| < 512$. However, $\log$ quantization with the $\mathrm{bitwidth}$ as linear results in a significantly lower number of quantization errors in the region $128<|\mathrm{LogQuant}(x)-x| < 512$. This comes at the expense of a slight increase in errors in the region $512<|\mathrm{LogQuant}(x)-x|$. Nonetheless, the quantization errors $\frac{1}{N}||\mathrm{LogQuant}(x)-x||_{1}=34.19$ for $\log$ and $\frac{1}{N}||\mathrm{LogQuant}(x)-x||_{1}=102.89$ for linear. 

\begin{figure}[ht]
\begin{center}
\centerline{\includegraphics[width=\columnwidth]{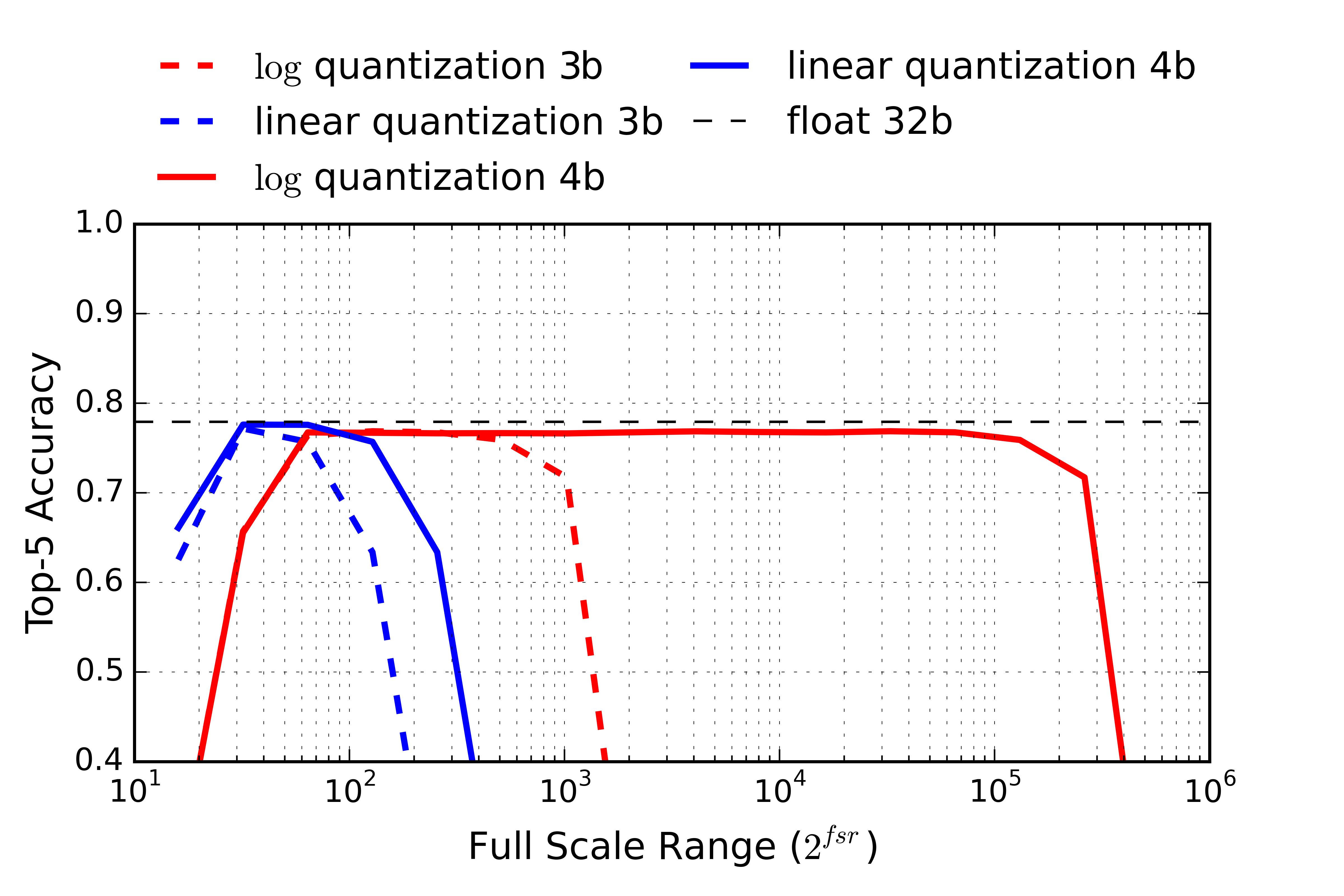}}
\caption{Top5 Accuracy vs Full scale range: AlexNet}
\label{fig:alex}
\end{center}
\end{figure} 

\begin{figure}[ht]
\begin{center}
\centerline{\includegraphics[width=\columnwidth]{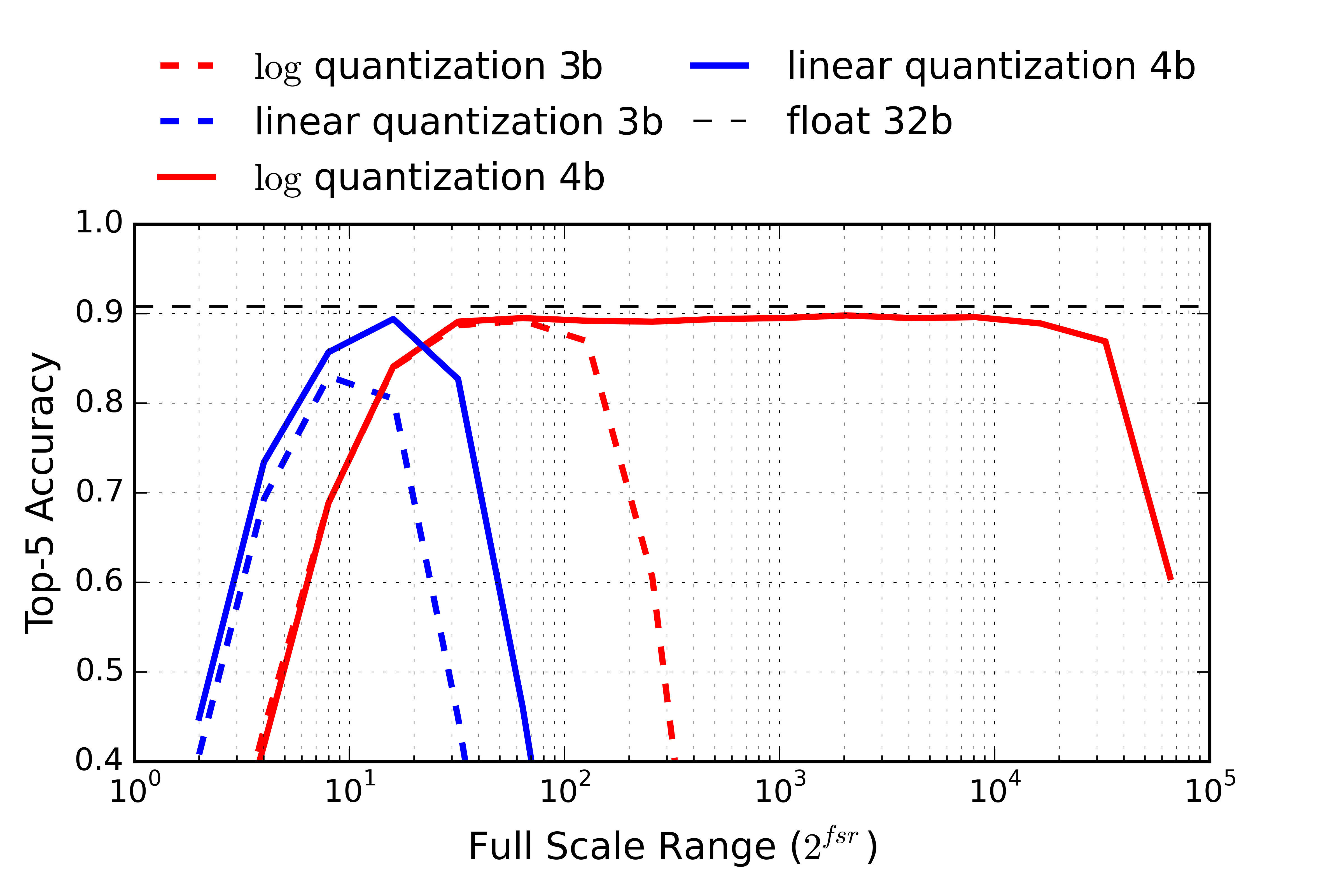}}
\caption{Top5 Accuracy vs Full scale range: VGG16}
\label{fig:vgg}
\end{center}
\end{figure} 

We run the models as described in Tables~\ref{table:alexnet} and ~\ref{table:vgg} and test on the validation set without data augmentation. We evaluate it with variable $\mathrm{bitwidth}$s and $\mathrm{FSR}$s for both quantizer layers. 

Figure~\ref{fig:alex} illustrates the results of AlexNet. Using only 3 bits to represent the activations for both logarithmic and linear quantizations, the top-5 accuracy is still very close to that of the original, unquantized model encoded at floating-point 32b. However, logarithmic representations tolerate a large dynamic range of $\mathrm{FSR}$s. For example, using 4b $\log$, we can obtain $3$ order of magnitude variations in the full scale without a significant loss of top-5 accuracy. We see similar results for VGG16 as shown in Figure~\ref{fig:vgg}. 
Table~\ref{table:act} lists the classification accuracies with the optimal $\mathrm{FSR}$s for each case. There are some interesting observations. First, $3$b $\log$ performs $0.2\%$ worse than $3$b linear for AlexNet but $6.2\%$ better for VGG16, which is a higher capacity network than AlexNet. Second, by encoding the activations in $3$b $\log$, we achieve the same top-5 accuracy compared to that achieved by $4$b linear for VGG16. Third, with $4$b $\log$, there is no loss in top-5 accuracy from the original float32 representation.

\begin{table}[H]
\caption{Top-5 accuracies with quantized activations at optimal $\mathrm{FSRs}$}
\label{table:act}
\vskip 0.15in
\begin{center}
\begin{small}
\begin{tabular}{lcc}
\hline
\abovespace\belowspace
Model & AlexNet & VGG16 \\
\hline
\abovespace
Float 32b & $78.3\%$ & $89.8\%$ \\
Log. 3b   & $76.9\%(\mathrm{fsr}=7)$ & $89.2\%(\mathrm{fsr}=6)$ \\
Log. 4b   & $76.9\%(\mathrm{fsr}=15)$ & $89.8\%(\mathrm{fsr}=11)$ \\
Linear 3b & $77.1\%(\mathrm{fsr}=5)$ & $83.0\%(\mathrm{fsr}=3)$ \\
\belowspace
Linear 4b & $77.6\%(\mathrm{fsr}=5)$ & $89.4\%(\mathrm{fsr}=4)$ \\
\hline
\end{tabular}
\end{small}
\end{center}
\vskip -0.1in
\end{table}

\subsection{Logarithmic Representation of Weights of Fully Connected Layers}
\label{sec:log quant of fc}

The FC weights are quantized using the same strategies as those in Section ~\ref{sec:log quant of act}, except that they have sign bit. We evaluate the classification performance using $\log$ data representation for both FC weights and activations jointly using method 2 in Section ~\ref{ssec:sub2}. For comparison, we use linear for FC weights and $\log$ for activations as reference. For both methods, we use optimal $4$b $\log$ for activations that were computed in Section ~\ref{sec:log quant of act}. 

Table~\ref{table_fc} compares the mentioned approaches along with floating point. We observe a small $0.4\%$ win for $\log$ over linear for AlexNet but a $0.2\%$ decrease for VGG16. Nonetheless, $\log$ computation is performed without the use of multipliers.

\begin{table}[H]
\caption{Top-5 accuracy after applying quantization to weights of FC layers}
\label{table_fc}
\vskip 0.15in
\begin{center}
\begin{small}
\begin{tabular}{lcccr}
\hline
\abovespace\belowspace
Model & Float 32b & Log. 4b & Linear 4b \\
\hline
\abovespace
AlexNet   & $76.9\%$ & $76.8\%$ & $76.4\%$ \\
\belowspace
VGG16     & $89.8\%$ & $89.5\%$ & $89.7\%$ \\
\hline
\end{tabular}
\end{small}
\end{center}
\vskip -0.1in
\end{table}

An added benefit to quantization is a reduction of the model size. By quantizing down to $4$b $\log$ including sign bit, we compress the FC weights for free significantly from $1.9$ Gb to $0.27$ Gb for AlexNet and $4.4$ Gb to $0.97$ Gb for VGG16. This is because the dense FC layers occupy $98.2\%$ and $89.4\%$ of the total model size for AlexNet and VGG16 respectively.

\subsection{Logarithmic Representation of Weights of Convolutional Layers}
\label{sec:log quant of conv}

We now represent the convolutional layers using the same procedure. We keep the representation of activations at $4$b $\log$ and the representation of weights of FC layers at $4$b $\log$, and compare our $\log$ method with the linear reference and ideal floating point. We also perform the dot products using two different bases: $2,\sqrt{2}$. Note that there is no additional overhead for $\log$ base-$\sqrt{2}$ as it is computed with the same equation shown in Equation \ref{eq:wandx3}.

 Table~\ref{table:conv} shows the classification results. The results illustrate an approximate $6\%$ drop in performance from floating point down to 5b base-2 but a relatively minor $1.7\%$ drop for 5b base-$\sqrt{2}$. They includes sign bit. There are also some important observations here. 

\begin{table}[H]
\caption{Top-5 accuracy after applying quantization to weights of convolutional layers}
\label{table:conv}
\vskip 0.15in
\begin{center}
\begin{small}
\begin{tabular}{lcccr}
\hline
\abovespace
Model & Float & Linear & Base-2 & Base-$\sqrt{2}$ \\
\belowspace
      & 32b & 5b &  Log 5b & Log 5b \\
\hline
\abovespace
AlexNet & $76.8\%$ & $73.6\%$ & $70.6\%$ & $75.1\%$ \\
\belowspace
VGG16   & $89.5\%$ & $85.1\%$ & $83.4\%$ & $89.0\%$ \\
\hline
\end{tabular}
\end{small}
\end{center}
\vskip -0.1in
\end{table}

We first observe that the weights of the convolutional layers for AlexNet and VGG16 are more sensitive to quantization than are FC weights. Each FC weight is used only once per image (batch size of 1) whereas convolutional weights are reused many times across the layer's input activation map. Because of this, the quantization error of each weight now influences the dot products across the entire activation volume. Second, we observe that by moving from $5$b base-$2$ to a finer granularity such as $5$b base-$\sqrt{2}$, we allow the network to 1) be robust to quantization errors and degradation in classification performance and 2) retain the practical features of $\log$-domain arithmetic. 

\begin{figure}[H]
\vskip 0.2in
\begin{center}
\centerline{\includegraphics[width=\columnwidth]{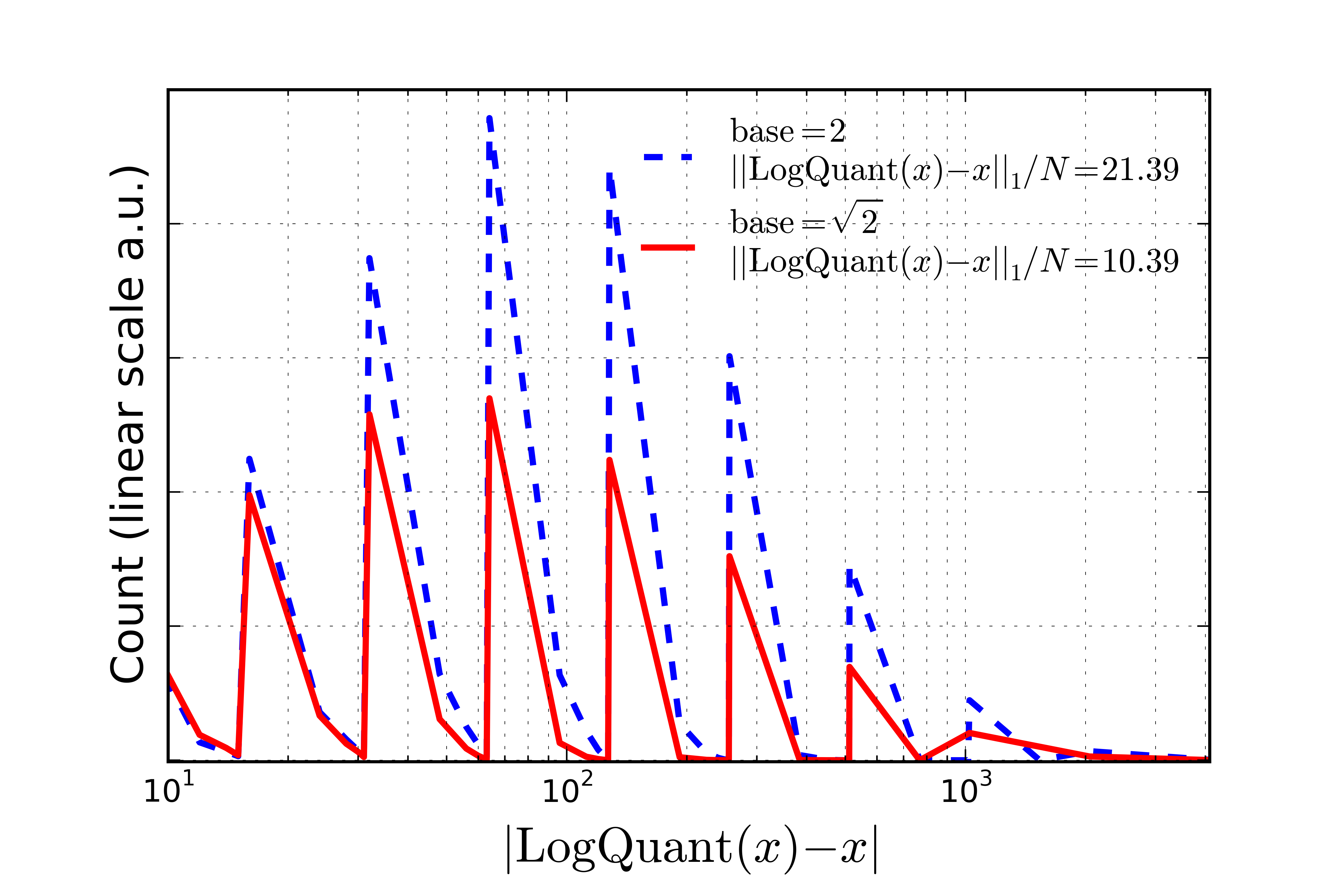}}
\caption{Distribution of quantization errors for weights under base 2 and $\sqrt{2}$.}
\label{fig:mse conv}
\end{center}
\vskip -0.2in
\end{figure} 

The distributions of quantization errors for both $5$b base-$2$ and $5$b base-$\sqrt{2}$ are shown in Figure ~\ref{fig:mse conv}. The total quantization error on the weights, $\frac{1}{N}||\mathrm{Quantize}(x)-x||_1$, where $x$ is the vectorized weights of size $N$, is $2\times$ smaller for base-$\sqrt{2}$ than for base-$2$.

\subsection{Training with Logarithmic Representation}
\label{sec:log quant of delta}
We incorporate $\log$ representation during the training phase. This entire algorithm can be computed using Method 2 in Section ~\ref{ssec:sub2}. Table~\ref{table:vgg_cifar10} illustrates the networks that we compare. The proposed $\log$ and linear networks are trained at the same resolution using 4-bit unsigned activations and $5$-bit signed weights and gradients using Algorithm ~\ref{alg:train} on the CIFAR10 dataset with simple data augmentation described in \cite{He15}. Note that unlike BinaryNet \cite{Courbariaux16}, we quantize the backpropagated gradients to train $\log$-net. This enables end-to-end training using logarithmic representation at the $5$-bit level. For linear quantization however, we found it necessary to keep the gradients in its unquantized floating-point precision form in order to achieve good convergence. Furthermore, we include the training curve for BinaryNet, which uses unquantized gradients. 

\begin{figure}[H]
\begin{center}
\centerline{\includegraphics[width=\columnwidth]{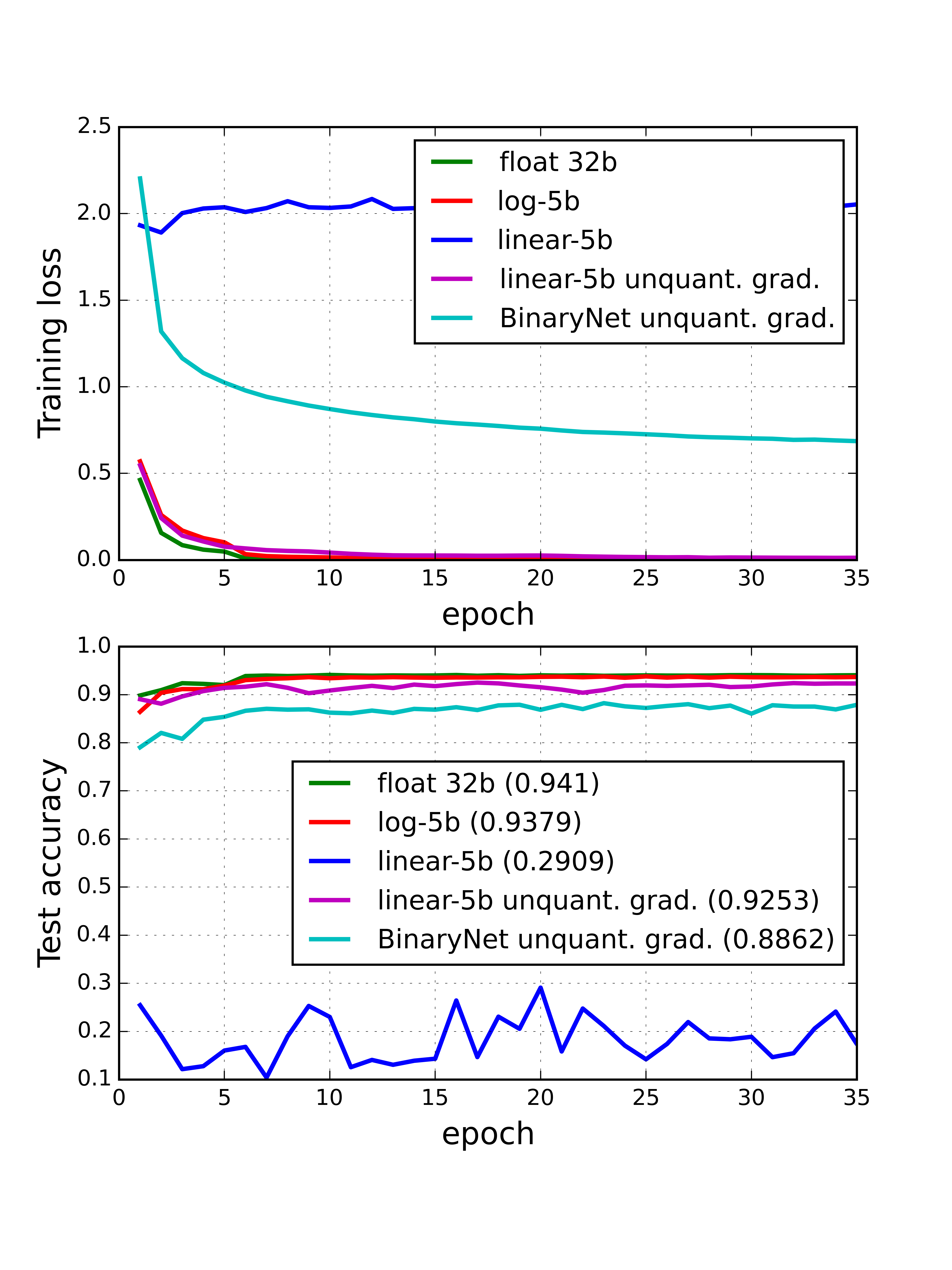}}
\caption{Loss curves and test accuracies}
\label{fig:training}
\end{center}
\end{figure}

Fig. ~\ref{fig:training} illustrates the training results of $\log$, linear, and BinaryNet. Final test accuracies for $\log$-$5$b, linear-$5$b, and BinaryNet are $0.9379$, $0.9253$, $0.8862$ respectively where linear-$5$b and BinaryNet use unquantized gradients. The test results indicate that even with quantized gradients, our proposed network with $\log$ representation still outperforms the others that use unquantized gradients. 
\begin{algorithm}[ht]
\begin{algorithmic}
    \REQUIRE a minibatch of inputs and targets $(a_0,a^*)$,
    previous weights $W$.
    \ENSURE updated weights $W^{t+1}$
    
    \STATE \COMMENT{1. Computing the parameters' gradient:}
    
    \STATE \COMMENT{1.1. Forward propagation:}
    \FOR{$k=1$ to $L$}  
        \STATE $W_k^q \leftarrow {\rm LogQuant}(W_k)$
        \STATE $a_k \leftarrow {\rm ReLU}\left(a_{k-1}^{q} W_{k}^{b}\right)$
        \STATE $a_k^q \leftarrow {\rm LogQuant}(a_k)$
    \ENDFOR
    
    \STATE \COMMENT{1.2. Backward propagation:}
    \STATE Compute $g_{a_L}=\frac{\partial C}{\partial a_L}$ knowing $a_L$ and $a^*$
    \FOR{$k=L$ to $1$} 
        \STATE $g_{a_{k}}^q \leftarrow {\rm LogQuant}(g_{a_{k}})$
        \STATE $g_{a_{k-1}} \leftarrow g_{a_{k}}^{q} W_k^{q}$
        \STATE $g_{W_{k}} \leftarrow g_{a_{k}}^{q\top} a_{k-1}^q$

    \ENDFOR
    
    \STATE \COMMENT{2. Accumulating the parameters' gradient:}
    \FOR{$k=1$ to $L$}
        \STATE $W_k^{t+1} \leftarrow {\rm Update}(W_k,g_{W_k})$
    \ENDFOR
\end{algorithmic}
\caption{Training a CNN with base-2 logarithmic representation. $C$ is the softmax loss for each minibatch. LogQuant(x) quantizes $x$ in base-2 $\log$-domain. The optimization step Update($W_k$,$g_{W_k}$) updates the weights $W_k$ based on backpropagated gradients $g_{W_k}$. We use the SGD with momentum and Adam rule.
}
\label{alg:train}
\end{algorithm}

\section{Conclusion}
In this paper, we describe a method to represent the weights and activations with low resolution in the $\log$-domain, which eliminates bulky digital multipliers. This method is also motivated by the non-uniform distributions of weights and activations, making $\log$ representation more robust to quantization as compared to linear. We evaluate our methods on the classification task of ILSVRC-2012 using pretrained models (AlexNet and VGG16). We also offer extensions that incorporate end-to-end training using $\log$ representation including gradients.

\begin{table}[H]
\caption{Structure of VGG-like network for CIFAR10}
\label{table:vgg_cifar10}
\begin{center}
\begin{small}
\begin{tabular}{ccc}
\hline
\abovespace\belowspace
$\log$ quantization & linear quantization & BinaryNet \\
\hline
\abovespace
Conv $64\cdot3\cdot3^2$ & Conv $64\cdot3\cdot3^2$ & Conv $64\cdot3\cdot3^2$\\
BatchNorm & BatchNorm & BatchNorm \\
ReLU & ReLU & - \\
{\bf LogQuant} & {\bf LinearQuant} & {\bf Binarize} \\
Conv $64\cdot64\cdot3^2$ & Conv $64\cdot64\cdot3^2$ & Conv $64\cdot64\cdot3^2$\\
BatchNorm & BatchNorm & BatchNorm \\
ReLU & ReLU & - \\
{\bf LogQuant} & {\bf LinearQuant} & {\bf Binarize} \\
MaxPool $2 \times 2$ & MaxPool $2 \times 2$ &MaxPool $2 \times 2$ \\ \hline
Conv $128\cdot64\cdot3^2$ & Conv $128\cdot64\cdot3^2$ & Conv $128\cdot64\cdot3^2$\\
BatchNorm & BatchNorm & BatchNorm \\
ReLU & ReLU & - \\
{\bf LogQuant} & {\bf LinearQuant} & {\bf Binarize} \\
Conv $128\cdot128\cdot3^2$ & Conv $128\cdot128\cdot3^2$ & Conv $128\cdot128\cdot3^2$\\
BatchNorm & BatchNorm & BatchNorm \\
ReLU & ReLU & - \\
{\bf LogQuant} & {\bf LinearQuant} & {\bf Binarize} \\
MaxPool $2 \times 2$ & MaxPool $2 \times 2$ &MaxPool $2 \times 2$ \\ \hline
Conv $256\cdot128\cdot3^2$ & Conv $256\cdot128\cdot3^2$ & Conv $256\cdot128\cdot3^2$\\
BatchNorm & BatchNorm & BatchNorm \\
ReLU & ReLU & - \\
{\bf LogQuant} & {\bf LinearQuant} & {\bf Binarize} \\
Conv $256\cdot256\cdot3^2$ & Conv $256\cdot256\cdot3^2$ & Conv $256\cdot256\cdot3^2$\\
BatchNorm & BatchNorm & BatchNorm \\
ReLU & ReLU & - \\
{\bf LogQuant} & {\bf LinearQuant} & {\bf Binarize} \\
Conv $256\cdot256\cdot3^2$ & Conv $256\cdot256\cdot3^2$ & Conv $256\cdot256\cdot3^2$\\
BatchNorm & BatchNorm & BatchNorm \\
ReLU & ReLU & - \\
{\bf LogQuant} & {\bf LinearQuant} & {\bf Binarize} \\
Conv $256\cdot256\cdot3^2$ & Conv $256\cdot256\cdot3^2$ & Conv $256\cdot256\cdot3^2$\\
BatchNorm & BatchNorm & BatchNorm \\
ReLU & ReLU & - \\
{\bf LogQuant} & {\bf LinearQuant} & {\bf Binarize} \\
MaxPool $2 \times 2$ & MaxPool $2 \times 2$ &MaxPool $2 \times 2$ \\ \hline
FC $1024\cdot256\cdot4^2$ & FC $1024\cdot256\cdot4^2$ & FC $1024\cdot256\cdot4^2$ \\
BatchNorm & BatchNorm & BatchNorm \\
ReLU & ReLU & - \\
{\bf LogQuant} & {\bf LinearQuant} & {\bf Binarize} \\ \hline
FC $1024\cdot1024$ & FC $1024\cdot1024$ & FC $1024\cdot1024$ \\
BatchNorm & BatchNorm & BatchNorm \\
ReLU & ReLU & - \\
{\bf LogQuant} & {\bf LinearQuant} & {\bf Binarize} \\ \hline
FC $10\cdot1024$ & FC $10\cdot1024$ & FC $10\cdot1024$ \\
- & - & BatchNorm \\
\hline
\end{tabular}
\end{small}
\end{center}
\end{table}

\bibliography{manuscript}
\bibliographystyle{icml2016}

\end{document}